# Enhanced Lung Cancer Survival Prediction using Semi-Supervised Pseudo-Labeling and Learning from Diverse PET/CT Datasets

Mohammad R. Salmanpour[1,2,3*], Arman Gorji[3,4], Amin Mousavi[3], Ali Fathi Jouzdani[3,4], Nima Sanati[3,4], Mehdi Maghsudi[5], Bonnie Leung[6], Cheryl Ho[1], Ren Yuan[2,6†], Arman Rahmim[1,2,7†]

[1] BC Cancer Research Institute, Vancouver, British Columbia (BC), Canada
[2] Department of Radiology, University of British Columbia, Vancouver, BC, Canada
[3] Technological Virtual Collaboration (TECVICO Corp.), Vancouver, BC, Canada
[4] NAIRG, Department of Neuroscience, Hamadan University of Medical Sciences, Hamadan, Iran
[5] Rajaie Cardiovascular Medical and Research Center, Iran University of Medical Science, Tehran, Iran
[6] BC Cancer, Vancouver Center, Vancouver, BC, Canada
[7] Department of Physics & Astronomy, University of British Columbia, Vancouver, BC, Canada

(\*) **Corresponding Author:**
Mohammad R. Salmanpour, PhD: *msalman@bccrc.ca*
Department of Integrative Oncology, BC Cancer Research Institute, Vancouver, BC V5Z 1L3, Canada, Tel: 604-675-8262.

(†) **Co-Last Authors:**
These authors had equal contributions as co-last authors.

## ABSTRACT

**Objective:** This study explores a semi-supervised learning (SSL), pseudo-labeled strategy using diverse datasets to enhance lung cancer (LCa) survival predictions, analyzing Handcrafted and Deep Radiomic Features (HRF/DRF) from PET/CT scans with Hybrid Machine Learning Systems (HMLS).
**Methods:** We collected 199 LCa patients with both PET & CT images, obtained from The Cancer Imaging Archive (TCIA) and our local database, alongside 408 head&neck cancer (HNCa) PET/CT images from TCIA. We extracted 215 HRFs and 1024 DRFs by PySERA and a 3D-Autoencoder, respectively, within the ViSERA software, from segmented primary tumors. The supervised strategy (SL) employed a HMLSs: PCA connected with 4 classifiers on both HRF and DRFs. SSL strategy expanded the datasets by adding 408 pseudo-labeled HNCa cases (labeled by Random Forest algorithm) to 199 LCa cases, using the same HMLSs techniques. Furthermore, Principal Component Analysis (PCA) linked with 4 survival prediction algorithms were utilized in survival hazard ratio analysis.
**Results:** SSL strategy outperformed SL method (p-value<0.05), achieving an average accuracy of 0.85±0.05 with DRFs from PET and PCA+ Multi-Layer Perceptron (MLP), compared to 0.65±0.08 for SL strategy using DRFs from CT and PCA+ K-Nearest Neighbor (KNN). Additionally, PCA linked with Component-wise Gradient Boosting Survival Analysis on both HRFs and DRFs, as extracted from CT, had an average c-index of 0.80 with a Log Rank p-value<<0.001, confirmed by external testing.
**Conclusions:** Shifting from HRFs and SL to DRFs and SSL strategies, particularly in contexts with limited data points, enabling CT or PET alone to significantly achieve high predictive performance.

## 1. INTRODUCTION

Lung cancer (LCa) is a global health issue, with 2.21 million new cases and over 1.8 million annual deaths projected by 2030, making it the second most diagnosed and leading cause of cancer deaths [1] [2] [3] [4]. Key challenges include the lack of effective biomarkers and delayed diagnosis, along with insufficient funding [5] [6]. Accurate prognostic models are needed to enhance clinical decision-making, with overall survival (OS) often used to assess treatment efficacy [7] [8] [9]. While supervised learning (SL) methods are effective, acquiring labeled data is costly and difficult [10] [11]. Semi-supervised learning (SSL), which uses limited labeled data with large unlabeled datasets, is valuable when labeled data is scarce [12] [13] [14]. SSL has shown promise in LCa prediction, outperforming fully supervised models [16], and has enhanced diagnostics in malignant diseases [17]. Recent studies using SSL on chest X-rays and CT scans have achieved state-of-the-art lung nodule detection [16].

Head and neck cancer (HNCa) and LCa share clinical and biological similarities, including common risk factors and genetic mutations, making HNCa data useful for improving LCa survival predictions [18]. Both cancers are linked to tobacco and alcohol use and exhibit comparable tumor microenvironments and treatment responses [19]. Survivors of HNCa face a higher risk of developing second primary cancers, including LCa, with 70%-80% of HNCas associated with prior tobacco use [20]. Additionally, both HNCa and LCa datasets include squamous cell carcinoma (SCC), while LCa datasets also feature non-small cell lung carcinoma (NSCLC) and adenocarcinoma (ADC), all of which share similar radiomic features [21] [22]. Incorporating HNCa data into an SSL framework allows us to enhance feature space representation, improving LCa outcome predictions. This connection between the two diseases supports using their images to improve OS predictions in LCa patients.

Positron emission tomography (PET) and computerized tomography (CT) are crucial diagnostic tools for detecting cancerous lesions and metabolic diseases. PET provides functional information about tissue metabolism, while CT offers detailed structural information [23] [24]. To employ machine learning (ML) algorithms in predictive performance, one may extract imaging features such as deep Radiomic features (DRF) and Handcrafted Radiomic features (HRF) [25] [26]. These quantitative features are valuable for disease diagnosis, therapy response, and prognosis prediction [27] [28]. HRFs are manually generated features, capturing known aspects like shape, intensity, and wavelet characteristics [29] [30], while DRFs are automatically learned through deep learning (DL) models, capturing more abstract patterns [31] [25]. Although DRFs provide deeper insights, HRFs offer greater interpretability and reproducibility [32] [33] [34] [35]. Combining DRFs and HRFs provides complementary information, enhancing image analysis [36].

A hybrid machine learning system (HMLS) combines different ML techniques to maximize their strengths and minimize limitations [37] [38] [39] [40]. HMLSs are widely used in medical diagnostics to improve prediction accuracy and reliability [41] [42] [43]. For example, integrating DL with Bayesian networks enhances prognosis forecasting for LCa patients [44] [45] [46]. DL extracts complex features from large datasets, while Bayesian networks incorporate prior knowledge and uncertainty, resulting in more accurate and interpretable outcomes [47]. HMLSs effectively address challenges such as missing data, noise, and diverse data sources in medical analysis [48]. Principal component analysis (PCA) reduces dimensionality in large datasets, improving interpretability and minimizing information loss by creating new uncorrelated variables (principal components) that maximize variance [49] [50]. This enhances model performance and generalizability, facilitating pattern identification and prediction accuracy, which this study aims to achieve.

Survival analysis is a statistical method used to predict the time until an event, such as disease onset or death, occurs [51]. It addresses time-to-event data, censored data, compares survival curves, and explores relationships between variables and survival time. Techniques like Cox proportional hazards regression allow for multiple covariates and generate hazard ratios for factors [52] [53] [54]. While some studies have examined survival analysis for outcome prediction [55] [56] [57], we propose a method that categorizes OS into two classes: class 1 for patients surviving over two years (the average survival time in LCa datasets) and class 2 for those deceased within two years. This approach, combined with survival regression algorithms (SRA), improves prediction accuracy and helps understand the relationship between features and outcomes for new patients.

This study explores the benefits of using diverse datasets, including HNCa, in an SSL approach with pseudo-labeling, alongside LCa datasets, to improve prediction performance compared to SL focused only on LCa. Key focuses include: i) addressing the challenge of creating large labeled datasets by utilizing SSL with limited labeled LCa data and a large amount of unlabeled HNCa data to predict OS in LCa, ii) comparing CT images with PET for their wider availability and lower cost, iii) comparing DRFs and HRFs for improved prediction accuracy, iv) evaluating SL using four binary-classification algorithms (CA) with PCA on HRF and DRF sets from 199 LCa PET/CT images, expanding with 408 pseudo-labeled HNCa cases, and v) investigating four hazard ratio survival analysis algorithms (SRA) with PCA for a deeper understanding of survival time and time-to-event patterns.

## 2. MATRIALS AND METHODS

### 2.1. Patient Data and Image Preprocessing

We utilized 199 LCa patients, obtained from The Cancer Imaging Archive (TCIA) and BC Cancer Agency (publicly funded, provincial comprehensive cancer care program that serves a population of 5.2 million residents in British Columbia, Canada.), alongside 408 HNCa with PET & CT images from TCIA. We only selected patients who had both CT and PET. We then selected patients that their PET images had AC (Attenuation Correction). AC is a process used to compensate for the loss of photons as they travel through the body [58]. These selection criteria yielded an LCa group of 33 from TCIA subjects (14 males, 19 females), 166 BC Cancer subjects (85 males, 81 females) and 408 HNCa from TCIA (320 males and 88 females). Supplemental Table S1 shows available demographics of these patients in different datasets. Binary OS was considered as outcome, categorizing them into two classes: class 1, alive after two years (averaged death time in LCa dataset), and class 2, deceased within two years of diagnosis. In the pre-processing step, PET images were first registered to CT. Subsequently, Standardized Uptake Value (SUV) correction was performed for raw PET data that enables standardized measurements of tracer uptake in medical imaging studies [59]. Moreover, the clipping technique scales lung cancer CT images by restricting intensity values to a range, typically from -300 to 300 Hounsfield units (HU). Finally, the minimum/maximum (min/max) normalization technique applied to both images to scale those [60]. Figure 1 shows that, unlike some studies [61] [62] that used automatic segmentation, this work relied on collaborative physicians to manually delineate cancer-affected areas on PET/CT images for imaging feature extraction.

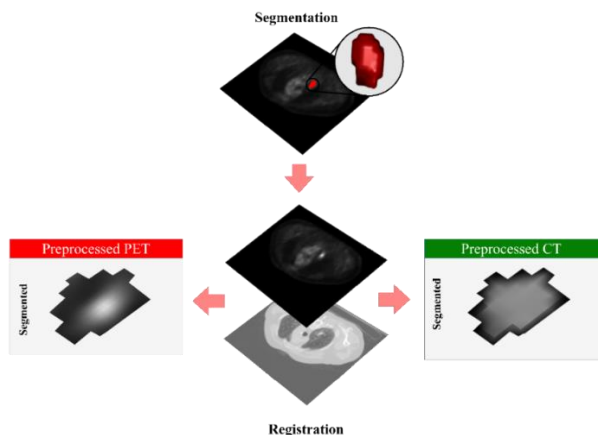

**Figure 1.** Example of PET, CT and segmented lung cancer area.

### 2.2. Study Procedure

The present study aimed to predict binary OS in patients with LCa, as shown in Figure 2. After image preprocessing and mask delineations (parts i and ii in Figure 2), as outlined in Section 2.1 and illustrated in Supplemental Figure S2, two imaging feature extraction methods—HRF and DRF—were employed for quantitative analysis (part iii). A total of 215 standardized HRFs were obtained using the standardized PySERA module in ViSERA software (*visera.ca*) [63]. Additional details on these HRFs can be found in Supplemental section 1.1. Similarly, 1024 DRFs were derived from the segmented PET and CT images from the bottleneck layer of the 3D Autoencoder embedded within the ViSERA software. The architecture of the Autoencoder is fully explained in Supplemental section 1.2. After DRF and HRF generation, 199 LCa datasets were split 80% for five-fold cross-validation and 20% for external nested testing (part iv). Additionally, stratification was performed to ensure that both divisions contained all outcomes. In addition, all extracted features were normalized by min-max function (part v). Moreover, different combinations of imaging feature sets such as DRF-CT, DRF-PET, HRF-CT, HRF-PET, DRF-CT plus HRF-CT, and DRF-CT plus HRF-CT were employed in this study.

For prediction, we employed two strategies—SL and SSL—within an HMLS framework, which integrated PCA with classification algorithms (CAs) or SRAs (part vi). By combining multiple approaches, the HMLS leverages the strengths of each method while mitigating their weaknesses [64] [65]. For the CAs, we used four classifiers: Multi-Layer Perceptron (MLP) [66], Support Vector Machine (SVM) [67], K-Nearest Neighbor (KNN) [68], and Ensemble Voting (EV) (part vii). For the SRAs, we utilized Fast Survival SVM (FSVM) [69], Component-wise Gradient Boosting Survival Analysis (CWGB) [70], Random Survival Forest (RSF) [71], and Cox Regression (COXR) (part viii) [72]. Hyperparameters for all classifiers were optimized using grid-search, as detailed in Supplemental Table S2, which significantly enhances the performance of ML algorithms. More detailed explanations of PCA, CAs, and RSAs are provided in Supplemental Sections 1.4 to 1.6.

SSL strategy incorporated a pseudo-labeling procedure, which utilizes both labeled and unlabeled data. Pseudo-labeling, as explained in Supplemental Section 1.7, involves training a model on labeled data to generate predictions for unlabeled data. These predictions, referred to as "pseudo labels," are then used to train the model on the previously unlabeled data in an SL manner. In this approach, a Random Forest (RandF) algorithm was employed to label the HNCa cases based on the training LCa data, and these pseudo-labeled HNCa cases were added to the training LCa dataset. During five-fold cross-validation, 80% of the 199 LCa datasets were divided into five folds. Four folds were used for training, while one-fold was used for validation. The RandF algorithm was trained on four training folds and used to label 408 HNCa datasets. The combined dataset of labeled HNCa samples and the four training folds was then input into four HMLS (PCA+CAs) models and tested on the remaining fold. Importantly, the validation and external testing sets were never used in the training or labeling process. SSL expanded the dataset by incorporating 408 pseudo-labeled HNCa cases alongside 199 LCa cases. In contrast, SL strategy used only the 199 labeled LCa PET and CT images, applying the same HMLS methods. We then compared the effectiveness of the expanded SSL with the SL methods.

For survival hazard ration analysis and prediction, we employed four HMLS (PCA+SRAs). Survival analysis used the median technique to categorize samples into low- and high-risk groups, training the SRAs with these categories alongside their continuous time data. A total of 199 LCa patients were included. Since SSL strategy is unsuitable for survival analysis due to the requirement of the last follow-up time, which is unpredictable, we exclusively used an SL strategy, relying on data from the 199 LCa patients. Therefore, OS data or last follow-up times were not available for HNCa patients in this study. Moreover, we utilized High Dimensional Hotelling's T Squared (HDTS) test for a further analysis of two groups including low- and high-risk patients [73]. Furthermore, we used mean function to determine high and low risk factor for patients, so the patients who had death time over average of death times were considered as low risk and patients lesser than death time average were considered as high risk. This method compares the meaning of two populations (high and low risk) when the feature number (F) of components is larger than the sample number (N), as elaborated in Supplemental Section 1.8. Thus, the HDTS test was applied to all LCa datasets including DRF-CT, DRF-PET, HRF-CT, and HRF-PET. Additionally, the average accuracy from five-fold cross-validation and external nested testing were calculated to evaluate the models in the classification task. The C-index and Log rank P-value were used to compare the survival risk assessments. Kaplan-Meier survival analysis [54] to examine the difference in OS between patients categorized as high risk or low risk by the model, and the differences between groups was assessed by the Log-Rank test [53].

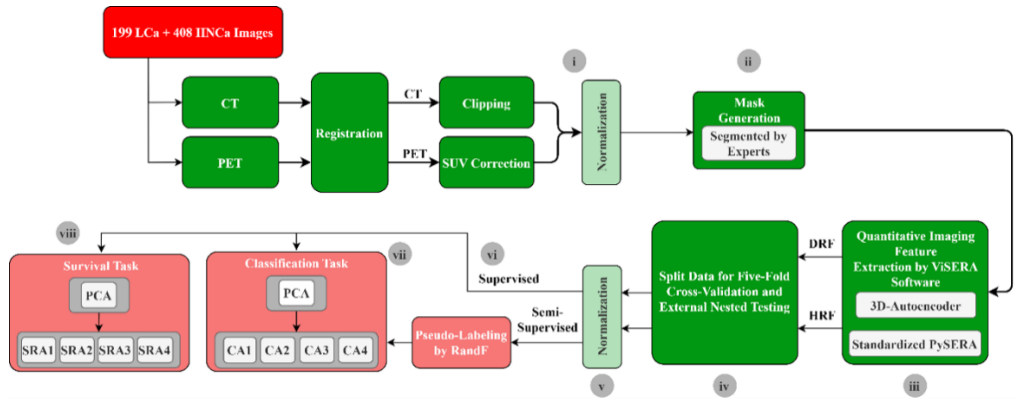

**Figure 2.** A schematic diagram of the proposed workflow. The study procedure includes i) in the image preprocessing step, ii) mask generation, iii) imaging feature extraction from the mask applied to preprocessed PET and CT images, iv) Splitting LCa dataset for five-fold cross-validation and external nested testing, v) normalizing HRFs and DRFs, vi) supervised and semi-supervised prediction tasks, vii) PCA linked with 4 CAs, viii) PCA linked with 4 SRA. SUV: Standardized Uptake Value, DRF: Deep Radiomic Feature, HRF: Handcrafted Radiomic Feature, PCA: Principal Component Analysis, CA: Classification Algorithm, SRA: Hazard Ratio Survival Analysis Algorithms, RandF: Random Forest Classifier, LCa: Lung Cancer, HNCa: Head and Neck Cancer.

## 3. RESULTS

This study investigated both SL and SSL strategies. Furthermore, two kinds of imaging features, namely HRFs and DRFs, were extracted from both PET and CT images. In addition, we used segmented masks to extract DRFs and HRFs from the images. Furthermore, different HMLSs are employed to predict OS through provided datasets using PCA combined with both CAs and SRAs.

### 3.1. Result Provided using Classification Analysis
### 3.1.1. Results Provided using HRF Sets:

In the HRF strategy, we first employed HRFs extracted from both PET and CT images as well as HMLSs mentioned in the method and martial section. Figure 3 illustrates that SL method incorporating HRF-PET, PCA, and KNN achieved the highest average accuracy, with a score of $0.62 \pm 0.04$ and an external nested test accuracy of $0.61 \pm 0.05$. Conversely, the strategy using HRF-CT, PCA, and SVM attained a slightly lower performance of $0.57 \pm 0.09$, but with a better external nested test result of $0.59 \pm 0.06$. Furthermore, the use of PCA, EV algorithms and HRF-PET, with an average accuracy of $0.58 \pm 0.05$ and an external nested test accuracy of $0.60 \pm 0.01$, does not add value to prediction tasks in an SL strategy.

In SSL strategy, the combination of HRF-PET, PCA, and MLP achieved the highest average accuracy of $0.77 \pm 0.10$ and an external nested test score of $0.72 \pm 0.02$. Meanwhile, using HRF-CT with the same HMLS setup resulted in an average accuracy of $0.75 \pm 0.06$ and an external nested test score of $0.68 \pm 0.03$. Additionally, other combinations, such as HRF-PET with PCA plus both SVM and KNN, achieved an average accuracy exceeding 0.76, while the combination of HRF-CT with PCA plus both SVM and KNN reached an average accuracy over 0.72 in SSL strategy. Furthermore, combining PCA, EV, and either HRF-CT or HRF-PET in an SSL, with average accuracies of $0.72 \pm 0.06$ and $0.73 \pm 0.07$ respectively, adds no value to the classification tasks. Additionally, these HMLSs achieved an external nested test accuracy of 0.68 ± 0.02. Overall, SSL methods using the HRF frameworks significantly outperformed, drawing on data from both PET and CT images (p-value < 0.05, paired t-test), compared to the top performance in SL strategy, which was 0.62 ± 0.04, achieved by a combination of HRF-PET, PCA, and KNN. All nested external testing performances are shown in Supplemental Table S3.

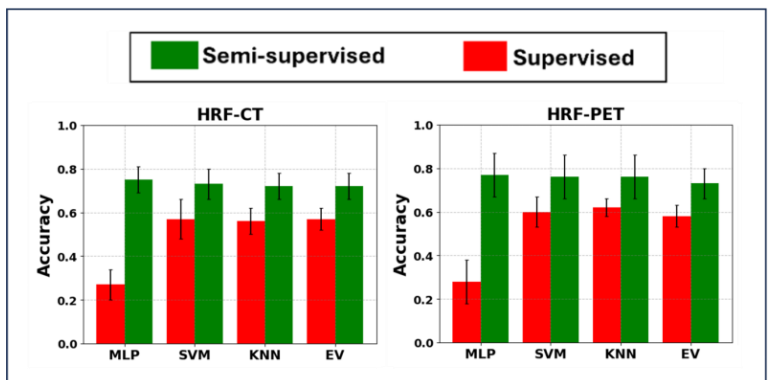

**Figure 3.** Bar plot of Mean ± standard deviation of Hybrid Machine Learning Systems (Principal Component Analysis linked with classifiers) while applied on (Left) HRF-CT: HRFs (handcrafted Radiomic features) extracted the segmented CT images and (Right) HRF-PET: HRFs extracted the segmented PET images. MLP: Multi-Layer Perceptron; SVM: Support Vector Machine, BR: Bagging Regression, KNN: K-Nearest Neighbor, EV: Ensemble Voting Algorithm.

### 3.1.2. Results Provided using DRF Sets:

Within the DRF framework, DRFs extracted from both PET and CT images were employed alongside the HMLSs outlined in the methods and materials section. Figure 4 shows that SL strategy using DRF-CT, PCA, and KNN achieved the highest average accuracy, registering at 0.65 ± 0.08, with an external nested test accuracy of 0.64 ± 0.06. On the other hand, the strategy utilizing DRF-PET with the same HMLS setup recorded a marginally lower average accuracy of 0.60 ± 0.05, but it resulted in a poorer external nested test score of 0.52 ± 0.04. Furthermore, the mixture of PCA, EV algorithms, and DRF-CT, with an average accuracy of 0.64 ± 0.03 and an external nested test accuracy of 0.65 ± 0.01, does not contribute to improving classification tasks in an SL strategy.

In SSL domain, the DRF-PET, PCA, and MLP configuration achieved the highest average accuracy, recording 0.85 ± 0.05 with an external nested test score of 0.80 ± 0.01. Similarly, the DRF-CT setup with the same HMLS configuration attained an average accuracy of 0.83 ± 0.06 and an external nested test accuracy of 0.79 ± 0.01. Additional SSL configurations, involving other HMLSs paired with DRF-PET or DRF-CT combined with PCA and both SVM and KNN, exceeded an average accuracy of 0.81. In summary, although the combination of DRF-PET, PCA, MLP, and SSL strategy achieved the highest average accuracy of 0.85, there is no significant difference in performance between SSL strategy combined with PCA and the three classifiers (MLP, SVM, and KNN) for both DRF-PET and DRF-CT (p-value > 0.05, paired t-test). This indicates that these algorithms within SSL strategy perform effectively. Collectively, SSL strategies leveraging the DRF framework significantly outperformed the highest SL result of 0.65 ± 0.08, achieved by DRF-CT, PCA, and KNN, as demonstrated by data from both PET and CT images and validated by a p-value < 0.05 from a paired t-test. Moreover, the superior performance of 0.85 ± 0.05 achieved by DRF-PET using an SSL strategy significantly surpassed the highest performance of 0.77 ± 0.06 achieved with HRF-PET when paired with PCA and MLP, as confirmed by a p-value < 0.05 in a paired t-test. All nested external testing performances are shown in Supplemental Table S3.

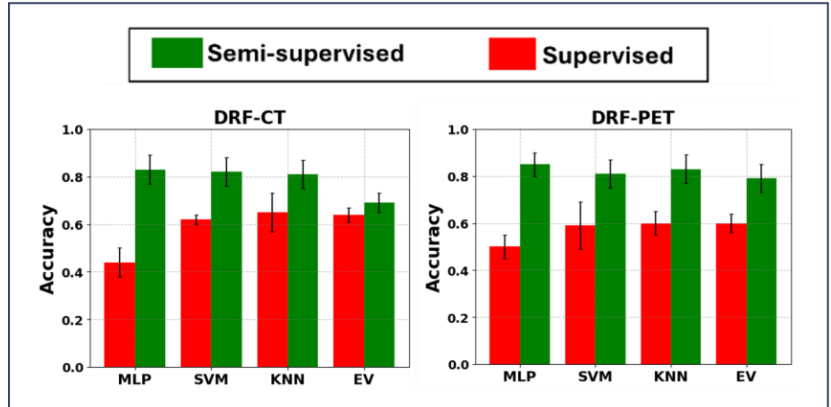

**Figure 4.** Bar plot of Mean ± standard deviation of Hybrid Machine Learning Systems (Principal Component Analysis linked with classifiers) while applied on (Left) DRF-CT: DRFs (deep Radiomic features) extracted the segmented CT images and (Right) DRF-PET: DRFs extracted the segmented PET images. MLP: Multi-Layer Perceptron; SVM: Support Vector Machine, BR: Bagging Regression, KNN: K-Nearest Neighbor, EV: Ensemble Voting Algorithm.

### 3.1.3. Results Provided using a Mixture of DRF and HRF Sets:

In SL strategy that combines HRF and DRF approaches, as illustrated in Figure 5, the incorporation of HRF-CT and DRF-CT with PCA and KNN yielded an average accuracy of 0.62 ± 0.09 and an external nested test accuracy of 0.57 ± 0.05. Additionally, the strategy using HRF-PET combined with DRF-PET, PCA, and KNN achieved a slightly lower average accuracy of 0.60 ± 0.07 but demonstrated a better external nested test accuracy of 0.59 ± 0.06. Furthermore, employing HRF-PET along with DRF-PET, PCA, and EV resulted in a comparable average accuracy of 0.60 ± 0.04, with an improved external nested test score of 0.62 ± 0.02. The differences in performance between these strategies are not statistically significant (p-value > 0.05, paired t-test).

In SSL strategy, the combination of HRF+DRF (extracted from PET), PCA, and MLP achieved the highest average accuracy of 0.78 ± 0.09 and an external nested test score of 0.75 ± 0.01. Additionally, other combinations such as HRF+ DRF (extracted from PET), PCA, and either SVM, KNN, or EV also achieved average accuracies exceeding 0.76, with no significant difference between these performances (p-value > 0.05, paired t-test). Similarly, SSL strategy using HRF-CT and DRF-CT with PCA and either MLP or SVM achieved an average accuracy of 0.74 ± 0.05 and an external nested test score above 0.70. Other combinations, such as the incorporation of HRF + DRF (extracted from CT) with either PCA+KNN or PCA+EV, achieved average accuracies exceeding 0.72, with no significant difference between these performances (p-value > 0.05, paired t-test). Overall, SSL strategy using HRF and DRF for both CT and PET significantly outperformed SL strategy (p-value < 0.05, paired t-test). All nested external test performances are detailed in Supplemental Table S3.

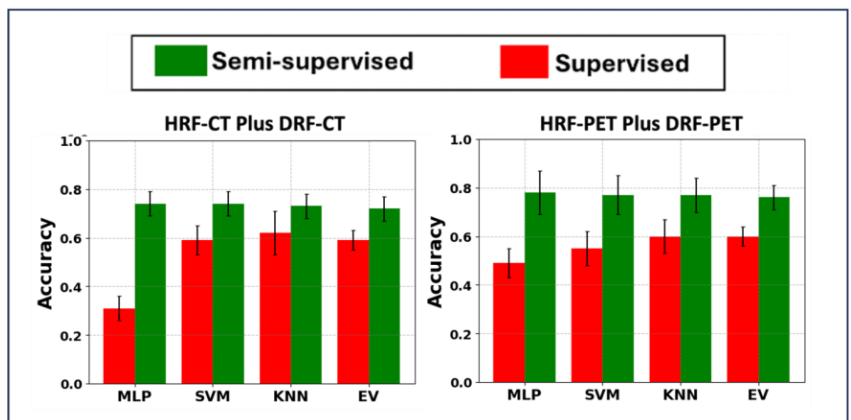

**Figure 5.** Bar plot of Mean ± standard deviation of Hybrid Machine Learning Systems (Principal Component Analysis linked with classifiers) while applied on (Left) Mixture of DRF-CT and HRF-CT; (Right) Mixture of DRF-PET and HRF-PET. DRF-CT: DRFs (deep Radiomic features) extracted the segmented CT images, HRF-CT: HRFs (handcrafted Radiomic features) extracted the segmented CT images, DRF-PET: DRFs extracted the segmented PET images. HRF-PET: HRFs extracted the segmented PET images. MLP: Multi-Layer Perceptron; SVM: Support Vector Machine, BR: Bagging Regression, KNN: K-Nearest Neighbor, EV: Ensemble Voting Algorithm.

## 3.2. Results Provided using Survival Analysis

The HDTS test indicated significant differences between DRFs extracted from CT and PET in both high-risk and low-risk groups, with P-values of 0.023 and 0.016, respectively. In contrast, there were no significant differences between HRFs extracted from CT and PET in the high-risk and low-risk groups, with P-values of 0.452 and 0.577, respectively. Subsequently, survival analysis was performed across six datasets: HRF-CT, HRF-PET, DRF-CT, DRF-PET, a mixture of DRF-PET & HRF-PET, and a mixture DRF-CT & HRF-CT using PCA combined with four SRAs. As depicted in Figure 6, the HRF frameworks utilizing PCA + CWGB on HRF-CT achieved an average C-Index of 0.79±0.08 and a Log Rank P-value significantly lesser than 0.001, significantly outperforming the same algorithms applied on HRF-PET, which had an average C-Index of 0.62±0.05 and a Log Rank P-value of 0.05 (p-value < 0.05, paired t-test). Additionally, these models recorded nested external testing C-Indexes of 0.80±0 and 0.59±0.03, respectively. In the DRF model, PCA + CWGB on DRF-CT resulted in an average C-Index of 0.80±0.10 and a Log Rank P-value significantly less than 0.001, significantly outperforming the same algorithms applied on DRF-PET, which had an average C-Index of 0.53±0.09 and a Log Rank P-value greater than 0.05 (p-value < 0.05, paired t-test). Furthermore, these models recorded nested external testing C-Indexes of 0.80±0 and 0.59±0.03, respectively. Meanwhile, there is no significant performance difference between the combinations of PCA and CWGB with either HRF-CT or DRF-CT (p-value > 0.05, paired t-test). Performances from PCA with FSVM or RSF on HRF-CT or DRF-PET were similar and below 0.65. All nested external testing performances are shown in Supplemental Table S4.

When prediction algorithms show similar but poor performance with DRFs and HRFs in both SL and SSL strategies, while others perform well, it suggests issues with algorithm compatibility, model complexity, or parameter tuning. Successful algorithms better utilize features and handle data noise, highlighting the need for optimization of the underperforming models. Figure 7 illustrates Kaplan-Meier survival curves for the most effective PCA + CWGB + HRFs and DRFs (extracted from CT images). Additionally, HRFs or DRFs derived from CT significantly outperformed those extracted from PET (p-value < 0.05, paired t-test). Moreover, additional Kaplan-Meier survival curves from other HMLSs are shown in Supplemental Figure S3.

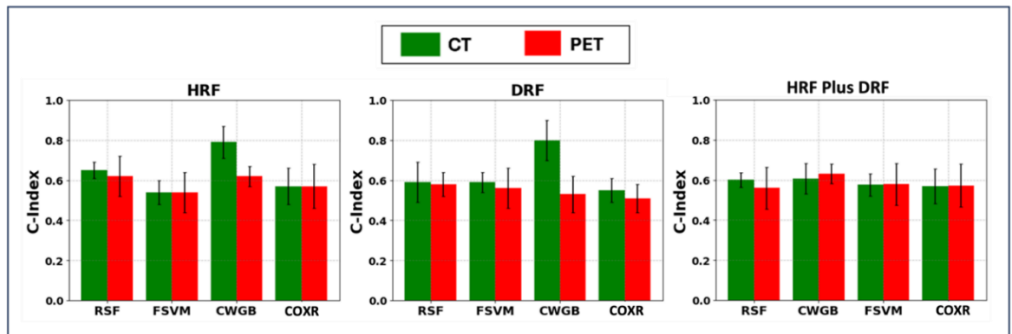

**Figure 6.** The performances provided from Hybrid Machine Learning Systems: Principal Component Analysis (PCA) linked with Fast Survival Support Vector Machine (FSVM), Component-wise Gradient Boosting Survival Analysis (CWGB), Random Survival Forest (RSF), and Cox Regression (COXR). 4 existing datasets included HRF (Handcrafted Radiomic Features) and DRFs (Deep Radiomic Features) extracted the segmented CT and PET images, respectively.

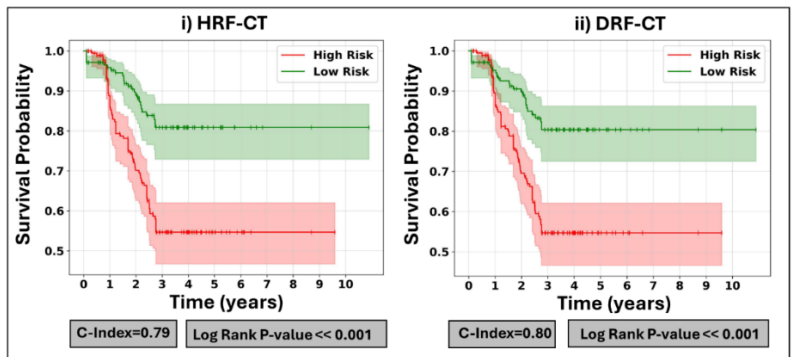

**Figure 7.** Variety of Kaplan-Meier survival curves generated by the top-performing Hybrid Machine Learning Systems (HMLS) including PCA + CWGB applied on (i) HRF-CT (HRF derived from CT images), and (ii) DRF-CT (DRF derived from CT images). PCA: Principal Component Analysis, HRF: Handcrafted Radiomic Features, DRF: Deep Radiomic Features, PCA: Principal Component Analysis, CWGB: Component-wise Gradient Boosting Survival Analysis.

## 4. DISCUSSION

LCa is the leading cause of cancer deaths globally, imposing significant economic burdens [74]. Accurate OS prediction, influenced by cancer type, stage, patient health, and treatment response [75] [76], is essential for treatment decisions, early diagnosis, and personalized therapy [77] [78] [79]. However, identifying reliable survival biomarkers remains challenging. Techniques like DL [79], radiomics [77], and gene expression signatures [75] [78] use clinical, imaging, and molecular data for prognostic models but face issues with data quality, missing values, and interpretability. Enhanced research is needed to improve OS prediction in LCa. SSL, which combines labeled and unlabeled data, is prevalent in NLP and bioinformatics but limited in medicine due to data quality and privacy concerns [30] [80] [81] [82] [83]. A review of 887 studies identified 169 prognostic factors for NSCLC, leading to a 12-factor survival model [84]. Additionally, an ensemble machine learning approach using metabolomic biomarkers from tumor biopsies accurately predicted patient survival, highlighting potential targets for improving outcomes [75].

Multiple ML approaches have been developed to predict OS in LCa [30]. Radiomics-based methods [30] extract features from CT images, enhancing survival prediction and biomarker identification. Wang et al. [79] and Zhao et al. [85] showed that Adaboost regression, combined with clinical, genomic, and gene expression data, outperforms other ML techniques in accuracy, precision, sensitivity, and specificity for predicting LCa survival and tumor stage. A study [86] found that Adaboost and Random Forest (RandF) models excel in predicting 5-year survival based on health-related quality of life. Dimension reduction techniques improve prediction accuracy and prevent overfitting by selecting fewer features [87] [88], with recent work achieving over 82% accuracy [89]. Stacked-ensemble meta-learners further enhance performance, achieving a 0.9 improvement in classifying short versus long survival times [75]. Advanced models include Yang et al.'s [90] long short-term memory DL using wearable activity data, Gupta et al.'s [91] ensemble of decision tree classifiers with clinical and demographic data, and Tajbakhsh et al.'s [92] semi-supervised multi-task learning on CT scans, which improved segmentation and reduced false positives.

In this study, we demonstrated that SSL strategy significantly outperformed SL approach in both HRF and DRF frameworks (p-value < 0.05, paired t-test). For example, HRF-PET combined with PCA and MLP in SSL method achieved an average accuracy of $0.77 \pm 0.10$, compared to $0.62 \pm 0.04$ using PCA and KNN in SL approach. Similarly, DRF-PET in SSL strategy with PCA and MLP achieved an average accuracy of $0.85 \pm 0.05$, outperforming the supervised DRF-CT approach, which achieved $0.65 \pm 0.08$. These findings show that DRF-PET in an SSL strategy offers the highest predictive accuracy. Additionally, DRF frameworks consistently outperformed HRF frameworks, with DRF-PET recording the highest accuracy of $0.85 \pm 0.05$, significantly better than HRF-PET at $0.77 \pm 0.10$.

HNCa and LCa share clinical and biological similarities, including tobacco and alcohol use, genetic mutations, similar tumor microenvironments, and treatment responses [18] [19]. Approximately 70–80% of HNCa cases are linked to tobacco use, increasing the risk of second primary cancers like LCa [20]. Both cancers frequently involve SCC with comparable radiomic features [21] [22]. Leveraging these similarities, HNCa datasets can augment LCa datasets within an SSL strategy. Although traditional SSL approaches with similar unlabeled data reduce the need for labeled large datasets, access to large amounts of these similar data, even unlabeled, might be limited or difficult. Our SSL strategy overcomes this limitation by incorporating diverse data, effectively addressing data size constraints across different classification tasks, as demonstrated in this study. Furthermore, DL networks like Autoencoders automatically extract complex and unknown features from raw data, whereas handcrafted radiomic methods rely on predefined features that may miss critical information requiring extensive domain knowledge. Consequently, DRF surpasses HRF in classification tasks by providing more comprehensive data representations and improved prediction performance. Importantly, in the HDTS tests, DRFs from PET and CT showed significant differences between high- and low-risk groups (P-values of 0.023 and 0.016, respectively), while no significant differences were observed for HRFs. This might indicate that CAs performed better with DRFs than with HRFs. By integrating SSL to minimize large data requirements with DRF extraction for comprehensive imaging analysis, our approach reduces reliance on costly and invasive modalities like PET, enabling CT alone to achieve high predictive performance. Furthermore, incorporating DLFs and HRFs did not enhance prediction performance compared to using DLFs or HRFs alone, as the distinct feature types can introduce redundant or conflicting information that may confuse the model and lead to overfitting.

In survival prediction tasks, the SSL approach was unsuitable due to the requirement of follow-up times, so we employed an SL using 199 LCa patients. Here, CT-based models using DRF and HRF frameworks showed superior performance compared to PET-based models (p-value < 0.05). The HRF framework achieved an average C-index of $0.79 \pm 0.08$, while the DRF framework showed a slightly higher C-index of $0.80 \pm 0.1$, both with highly significant Log-Rank P-values below 0.001. Given our comparison studies between DRFs and HRFs, further investigation into the utilization of DRFs as an alternative to HRFs could potentially provide significant enhancements in performance and robustness [93] [94].

Despite promising results, our study has several limitations. The datasets were restricted to specific cancer types (LCa and HNCa) and imaging modalities (CT and PET), necessitating validation with more diverse datasets. SSL strategy, reliant on pseudo-labeling, may introduce noise, emphasizing the need for high-quality pseudo-labels. Additionally, datasets with different diseases but identical imaging modalities could be explored. While DRFs capture complex patterns, they are less interpretable than HRFs and combining them with improved performance but reduced interpretability. The limited sample size may affect generalizability, requiring further validation on independent datasets. Although feature selection is a future area of interest, we used PCA to reduce feature size and prevent overfitting. For survival predictions, SSL strategy was unfeasible due to the need for the last follow-up date, which cannot be predicted. In future studies, applying SSL to survival analysis could be valuable if the latest follow-up times or OS data from diverse patient datasets become available.

## 5. CONCLUSION

This study applied semi-supervised ML with pseudo-labels and compared it to a supervised-only strategy for predicting OS. Key findings include: i) both CT and PET images performed well in SSL strategies within the DRF framework, achieving accuracies of 0.83±0.06 and 0.85±0.05, respectively, using PCA and MLP; ii) DRF frameworks outperformed HRF frameworks, with DRF achieving 0.85±0.05 versus HRF's 0.77±0.10; also, combining DRF and HRF did not improve OS prediction beyond DRF-PET in SSL strategy, which achieved 0.85±0.05. iii) SSL strategy (PCA, MLP, DRF-PET) significantly outperformed SL strategy (PCA, KNN, DRF-CT) with an accuracy of 0.85±0.05 versus 0.65±0.08. This study showed that integrating SSL with DRFs reduces dependence on costly modalities like PET, enabling CT alone to achieve high predictive performance. In survival analysis, HRF and DRF frameworks using CT images and PCA + CWGB showed superior performance compared to PET, achieving a C-index of ~0.80 with highly significant Log-Rank P-values (<0.001).

## DATA AND CODE AVAILIILITY
All code (including predictor and dimension reduction algorithms) is publicly shared at:
*https://github.com/MohammadRSalmanpour/Semi-Supervised-approach-for-servival-prediction-in-Lung-cancer-/tree/main*

## CONFILICT OF INTREST
The authors have no relevant conflicts of interest to disclose.

## ACHNOWLEDGEMNTS
This study was supported by the Natural Sciences and Engineering Research Council of Canada (NSERC) Discovery Grant RGPIN-2019-06467, UBC Department of Radiology 2023 AI Fund, as well as the Technological Virtual Collaboration Corporation (TECVICO CORP.), Vancouver, Canada.

## REFERENCES

[1] A. J. Alberg, M. V. Brock and et al, "Epidemiology of lung cancer: Diagnosis and management of lung cancer, 3rd ed: American College of Chest Physicians evidence-based clinical practice guidelines," *Chest,* vol. 143, no. 5, pp. e1S-e29S, May 2013.

[2] J. Ferlay, M. Colombet and et al, "cancer statistics for the year 2020: An overview," *Int J Cancer,* vol. 149, no. 4, pp. 778-789, Apr 5 2021.

[3] O. Menyhárt, J. T. Fekete and B. Győrffy, "Demographic shift disproportionately increases cancer burden in an aging nation: current and expected incidence and mortality in Hungary up to 2030," *Clin Epidemiol,* vol. 10, pp. 1093-1108, 2018.

[4] A. K. Ganti, A. B. Klein and et al, "Update of Incidence, Prevalence, Survival, and Initial Treatment in Patients With Non–Small Cell Lung Cancer in the US," *JAMA Oncol,* vol. 7, no. 12, pp. 1824-1832, 2021.

[5] F. R. Hirsch, G. V. Scagliotti and et al, "Lung cancer: current therapies and new targeted treatments," *Lancet,* vol. 389, no. 10066, pp. 299-311, 2017.

[6] R. Sullivan, O. I. Alatise and et al, "Global cancer surgery: delivering safe, affordable, and timely cancer surgery," *The Lancet. Oncology,* vol. 16, no. 11, pp. 1193-1224, Sep 2015.

[7] S. Doppalapudi, R. G. Qiu and Y. Badr, "Lung cancer survival period prediction and understanding: Deep learning approaches," *International Journal of Medical Informatics,* vol. 148, p. 104371, 2021.

[8] P. K. Cheema and R. Burkes, "Overall survival should be the primary endpoint in clinical trials for advanced non-small-cell lung cancer," *Curr Oncol,* vol. 20, no. 2, p. e150–e160, Apr 2013.

[9] R. Nooreldeen and H. Bach, "Current and Future Development in Lung Cancer Diagnosis," *Int J Mol Sci,* vol. 22, no. 16, p. 8661, Aug 12 2021.

[10] Z. Zhou, "A brief introduction to weakly supervised learning," *National Science Review,* vol. 5, no. 1, p. 44–53, 2018.

[11] M. J. Willemink, W. A. Koszek and et al, "Preparing Medical Imaging Data for Machine Learning," *Radiology,* vol. 295, no. 1, pp. 4-15, 2020.

[12] C. Ge, I. Gu and et al, "Deep semi-supervised learning for brain tumor classification," *BMC Medical Imaging,* vol. 20, no. 87, pp. 1-11, 2020.

[13] I. Triguero, S. García and F. Herrera , "Self-labeled techniques for semi-supervised learning: taxonomy, software and empirical study," *Knowl Inf Syst,* vol. 42, p. 245–284, 2015.

[14] P. Cunningham, M. Cord and S. J. Delany, "Supervised Learning," in *Machine Learning Techniques for Multimedia: Case Studies on Organization and Retrieval*, Berlin, Heidelberg, Springer Berlin Heidelberg, 2008, pp. 21-49.

[15] T. Huynh, A. Nibali and Z. He, "Semi-supervised learning for medical image classification using imbalanced training data," *Comput Methods Programs Biomed,* vol. 216, p. 106628, 2022.

[16] K. Shak, M. Al-Shabi and et al, "A new semi-supervised self-training method for lung cancer prediction," *arXiv preprint arXiv:2012.09472,* 2020.

[17] J. N. Eckardt, M. Bornhäuser and et al, "Semi-supervised learning in cancer diagnostics," *Front Oncol,* vol. 12, p. 960984, 2022.

[18] J. Malhotra, M. Malvezzi and et al, "Risk factors for lung cancer worldwide," *Eur Respir J,* vol. 48, no. 3, pp. 889-902, 2016.

[19] A. Jethwa and S. Khariwala , "Tobacco-related carcinogenesis in head and neck cancer," *Cancer and Metastasis Reviews,* vol. 36, pp. 411-423, 2017.

[20] J. Cramer , J. Grauer and et al, "Incidence of second primary lung cancer after low-dose computed tomography vs chest radiography screening in survivors of head and neck cancer: a secondary analysis of a randomized clinical trial," *JAMA Otolaryngology–Head & Neck Surgery,* vol. 147, no. 12, pp. 1071-8, 2021.

[21] S. Larsson , P. Carter and et al, "A mendelian randomisation study in UK Biobank and international genetic consortia participants," *PLoS medicine,* vol. 17, no. 7, p. e1003178, 2020.

[22] M. Vallieres , E. Kay-Rivest and et al, "Radiomics strategies for risk assessment of tumour failure in head-and-neck cancer.," *Scientific reports,* vol. 7, no. 1, p. 10117, 2017.

[23] L. Filippi and O. Schillaci, "Total-body [18F]FDG PET/CT scan has stepped into the arena: the faster, the better. Is it always true?," *Eur J Nucl Med Mol Imaging,* vol. 49, no. 10, pp. 3322-3327, Aug 2022.

[24] Z. Chen, X. Chen and R. Wang, "Application of SPECT and PET / CT with computer-aided diagnosis in bone metastasis of prostate cancer: a review," *Cancer Imaging,* vol. 22, no. 1, p. 18, Apr 2022.

[25] M. R. Salmanpour, S. M. Rezaeijo and et al, "Deep versus Handcrafted Tensor Radiomics Features: Prediction of Survival in Head and Neck Cancer Using Machine Learning and Fusion Techniques," *Diagnostics,* vol. 13, no. 10, p. 1696, 2023.

[26] M. Hosseinzadeh, A. Gorji and et al, "Prediction of Cognitive Decline in Parkinson's Disease Using Clinical and DAT SPECT Imaging Features, and Hybrid Machine Learning Systems," *Diagnostics,* vol. 13, no. 10, p. 1691, 2023.

[27] A. Zwanenburg, M. Vallières and et al, "The image biomarker standardization initiative: Standardized quantitative radiomics for high-throughput image-based phenotyping," *Radiology,* vol. 295, no. 2, pp. 328-338, 5 2020.

[28] M. Salmanpour, M. Ramezani and et al, "Deep versus handcrafted tensor radiomics features: application to survival prediction in head and neck cancer," in *EANM*, 2022.

[29] M. R. Salmanpour, M. Hosseinzadeh and e. al, "ViSERA: Visualized & Standardized Environment for Radiomics Analysis," *Society of Nuclear Medicine,* vol. 64, no. supplement 1, pp. P1131-P1131, 2022.

[30] X. Zhang, Y. Zhang and et al, "Deep Learning With Radiomics for Disease Diagnosis and Treatment: Challenges and Potential," *Front Oncol,* vol. 12:773840, p. 773840, 17 Feb 2022.

[31] M. Astaraki , C. Wang and et al, "A Comparative Study of Radiomics and Deep-Learning Based Methods for Pulmonary Nodule Malignancy Prediction in Low Dose CT Images," *Frontiers in Oncology,* vol. 11, p. 737368, 2021.

[32] M. W. Wagner, K. Namdar and et al, "Radiomics, machine learning, and artificial intelligence—what the neuroradiologist needs to know," *Neuroradiology,* vol. 63, no. 12, pp. 1957-1967, 2021.

[33] P. Afshar, A. Mohammadi and et al, "From Handcrafted to Deep-Learning-Based Cancer Radiomics: Challenges and Opportunities," *IEEE Signal Processing Magazine,* vol. 36, no. 4, pp. 132-160, July 2019.

[34] M. R. Salmanpour, M. Hosseinzadeh and et al, "Robustness and Reproducibility of Radiomics Features from Fusions of PET-CT Images," *J. Nuclear Med,* vol. 63, no. supplement 2, p. 3179, 2022.

[35] M. R. Salmanpour , M. Hosseinzadeh and et al, "Reliable and Reproducible Tensor Radiomics Features in Prediction of Survival in Head and Neck Cancer," in *EANM*, 2022.

[36] A. Bizzego, N. Bussola and et al, "Integrating deep and radiomics features in cancer bioimaging," in *IEEE Conference on Computational Intelligence in Bioinformatics and Computational Biology (CIBCB)*, Siena, Italy, 2019.

[37] H. Hefny, A. Abdel-Wahab and et al, "A Novel Framework for Hybrid Intelligent Systems," in *Multiple Approaches to Intelligent Systems*, Berlin, Heidelberg, 1999.

[38] P. Sornsuwit and S. Jaiyen, "A New Hybrid Machine Learning for Cybersecurity Threat Detection Based on Adaptive Boosting," *Applied Artificial Intelligence,* vol. 33, no. 5, pp. 462-482, 2019.

[39] M. Salmanpour, G. Hajianfar and et al, "Multitask outcome prediction using hybrid machine learning and PET-CT fusion radiomics," in *Journal of Nuclear Medicine*, 2021.

[40] M. Salmanpour, M. Shamsaei and et al, "Machine learning methods for optimal prediction of outcome in Parkinson's disease," in *n 2018 IEEE Nuclear Science Symposium and Medical Imaging Conference Proceedings (NSS/MIC)*, 2018.

[41] I. Kononenko, "Machine learning for medical diagnosis: history, state of the art and perspective," *Artificial Intelligence in medicine,* vol. 23, no. 1, pp. 89-109, 2001.

[42] M. R. Salmanpour, A. Saberi and et al, "Optimal feature selection and machine learning for prediction of outcome in Parkinson's disease," *Journal of Nuclear Medicine,* vol. 61, no. supplement 1, pp. 524-524, 2020.

[43] M. R. Salmanpour, M. Hosseinzadeh and et al, "Prediction of TNM stage in head and neck cancer using hybrid machine learning systems and radiomics features," *Medical Imaging 2022: Computer-Aided Diagnosis (spie),* vol. 12033, no. sUPPLEMENT 1, pp. 648-653, 2022.

[44] M. R. Salmanpour, M. Hosseinzadeh and et al, "Tensor Deep versus Radiomics Features: Lung Cancer Outcome Prediction using Hybrid Machine Learning Systems," in *Journal of Nuclear Medicine*, 2023.

[45] A. Gorji, A. F. Jouzdani and et al, "PET-CT Fusion Based Outcome Prediction in Lung Cancer using Deep and Handcrafted Radiomics Features and Machine Learning," in *Journal of Nuclear Medicine*, 2023.

[46] N. A. Othman, M. A. Abdel-Fattah and A. T. Ali, "A Hybrid Deep Learning Framework with Decision-Level Fusion for Breast Cancer Survival Prediction," *Big Data and Cognitive Computing,* vol. 7, no. 1, p. 50, 2023.

[47] D. Koller and N. Friedman, Probabilistic Graphical Models: Principles and Techniques - Adaptive Computation and Machine Learning, The MIT Press, 2009.

[48] R. Bellazzi and B. Zupan, "Predictive data mining in clinical medicine: current issues and guidelines," *International journal of medical informatics,* vol. 77, no. 2, pp. 81-97, 2008.

[49] I. Jolliffe and J. Cadima, "Principal component analysis: a review and recent development," *Philosophical transactions of the royal society A: Mathematical, Physical and Engineering Sciences,* vol. 374, no. 2065, p. 20150202, 2016.

[50] B. Hasan and A. Abdulazeez, "A review of principal component analysis algorithm for dimensionality reduction," *Journal of Soft Computing and Data Mining,* vol. 2, no. 1, pp. 20-30, 2021.

[51] S. Rai, P. Mishra and U. Ghoshal, "Survival analysis: A primer for the clinician scientists," *Indian Journal of Gastroenterology,* vol. 40, p. 541–549, 2021.

[52] S. V. Deo, V. Deo and V. Sundaram, "Survival analysis—part 2: Cox proportional hazards model," *Indian J Thorac Cardiovasc Surg,* vol. 37, no. 2, pp. 229-233, 2021.

[53] L. Cui, H. Li and et al, "A deep learning-based framework for lung cancer survival analysis with biomarker interpretation," *BMC Bioinformatics,* vol. 21, no. 1, p. 112, 2020.

[54] M. K. Goel, P. Khanna and J. Kishore, "Understanding survival analysis: Kaplan-Meier estimate," *Int J Ayurveda Res,* vol. 1, no. 4, pp. 274-8, 2010.

[55] O. P. Kolesnik, A. I. Shevchenko and et al, "Usage of Cox-Regression Model for Forecasting of Survival Rate in Patients with the Early Stage of Non-Small Cell Lung Cancer," *Advances in Lung Cancer,* vol. 3, no. 1, p. 26–33, 2014.

[56] Lavanya C, Pooja S and et al, "Novel Biomarker Prediction for Lung Cancer Using Random Forest Classifiers," *Cancer Inform,* vol. 22:11769351231167992, p. 11769351231167992, 21 Apr 2023.

[57] R. Al Mamlook, "Lung Cancer Survival Prediction Using Random Forest Based Decision Tree Algorithms," in *Proceedings of the International Conference on Industrial Engineering and Operations Management*, Washington, DC, 2018.

[58] D. Izquierdo-Garcia and C. Catana, "MR Imaging-Guided Attenuation Correction of PET Data in PET/MR Imaging," *PET Clin,* vol. 11, no. 2, pp. 129-149, Apr 2016.

[59] G Lucignani, G Paganelli and E Bombardieri, "The use of standardized uptake values for assessing FDG uptake with PET in oncology: a clinical perspective," *Nucl Med Commun,* vol. 25, no. 7, pp. 651-656, Jul 2004.

[60] A. Brahim, J.M. Górriz and et al, "Intensity normalization of DaTSCAN SPECT imaging using a model-based clustering approach," *Applied Soft Computing,,* vol. 37, pp. 234-244, 2015.

[61] S. Roy, D. Bhattacharyya and et al, "An Iterative Implementation of Level Set for Precise Segmentation of Brain Tissues and Abnormality Detection from MR Images," *IETE Journal of Research,* vol. 63, no. 6, pp. 769-783, 2017.

[62] M. R. Salmanpour, G. Hajianfar and et al, "Advanced Automatic Segmentation of Tumors and Survival Prediction in Head and Neck Cancer," in *Head and Neck Tumor Segmentation and Outcome Prediction*, Cham, 2021.

[63] M. R. Salmanpour, I. Shiri and et al, "ViSERA: Visualized & Standardized Environment for Radiomics Analysis - A Shareable, Executable, and Reproducible Workflow Generator," in *Proc. IEEE Medical Imaging Conference, 2023*, Vancouver, 2023.

[64] G. Hajianfar, K. Samira and et al, "Prediction of Parkinson's disease pathogenic variants using hybrid Machine learning systems and radiomic features," *Phys Med,* vol. 113, p. 102647, 2023 Aug 12.

[65] M. R. Salmanpour, M. Hosseinzadeh and e. al, "Prediction of drug amount in Parkinson's disease using hybrid machine learning systems and radiomics features," *International Journal of Imaging Systems and Technology,* vol. 33, no. 4, pp. 1437-1449, 2023.

[66] G. Cybenko, "Approximation by superpositions of a sigmoidal function," *Math. Control Signal Systems,* vol. 2, no. 4, p. 303–314, 1989.

[67] C. Cortes and V. Vapnik, "Support-vector networks," *Machine Learning.,* vol. 20, no. 3, p. 273–297, 1995.

[68] L. Peterson, "K-nearest neighbor," *Scholarpedia,* vol. 4, no. 2, p. 1883, 2009.

[69] S. Pölsterl, "A Library for Time-to-Event Analysis Built on Top of scikit-learn," *Journal of Machine Learning Research,* vol. 21, no. 212, p. 1–6, 2020.

[70] S. Pölsterl, "Gradient Boosted Models," 2020. [Online]. Available: https://scikit-survival.readthedocs.io/en/stable/user_guide/boosting.html. [Accessed 9 January 2024].

[71] S. Pölsterl, "Random survival forests," 2020. [Online]. Available: https://scikit-survival.readthedocs.io/en/stable/user_guide/random-survival-forest.html. [Accessed 9 January 2024].

[72] M. Du, D. Haag and et al, "Comparison of the tree-based machine learning algorithms to Cox regression in predicting the survival of oral and pharyngeal cancers: analyses based on SEER database," *Cancers,* vol. 12, no. 10, p. 2802, 2020.

[73] P. Secchi, A. Stamm and S. Vantini, "Inference for the mean of large p small n data: A finite-sample high-dimensional generalization of Hotelling's theorem," *Electronic Journal of Statistics,* vol. 7, pp. 2005-2031, 2013.

[74] R. Lewis, M. Hendry and et al, "Pragmatic methods for reviewing exceptionally large bodies of evidence: systematic mapping review and overview of systematic reviews using lung cancer survival as an exemplar," *Systematic Reviews,* vol. 8, no. 1, p. 171, 16 July 2019.

[75] H. A. Miller, V. H. van Berkel and H. B. Frieboes, "Lung cancer survival prediction and biomarker identification with an ensemble machine learning analysis of tumor core biopsy metabolomic data," *Metabolomics,* vol. 18, no. 8, p. 57, 2022.

[76] "Canadian Cancer Society. Prognosis and survival for lung cancer," May 2020. [Online]. Available: https://cancer.ca/en/cancer-information/cancer-types/lung/prognosis-and-survival. [Accessed 2 September 2023].

[77] N. C. DAmico, R. Sicilia and et al, "Radiomics-Based Prediction of Overall Survival in Lung Cancer Using Different Volumes-Of-Interest," *Applied Sciences,* vol. 10, no. 18, p. 6425, 2020.

[78] A. A. Dherasi, T. Q. Huang and et al, "A seven-gene prognostic signature predicts overall survival of patients with lung adenocarcinoma (LUAD)," *Cancer Cell Int,* vol. 21, no. 1, p. 294, 2021.

[79] S. Wang, H. Zhang and et al, "A Novel Deep Learning Method to Predict Lung Cancer Long-Term Survival With Biological Knowledge Incorporated Gene Expression Images and Clinical Data," *Front Genet,* vol. 13, pp. 800-853, 2022.

[80] J. E. van Engelen and H. Hoos, "A survey on semi-supervised learning," *Machine Learning,* vol. 109, no. 2, pp. 373 - 440, 2019.

[81] R. Shams, "Semi-supervised classification for natural language processing," *arXiv preprint arXiv:1409.7612,* 25 Sep 2014.

[82] M. Chen, Y. Hao and et al, "Disease prediction by machine learning over big data from healthcare communities," *IEEE Access,* vol. 5, pp. 8869-8879, 2017.

[83] A. Aliper, S. Plis and et al, "Deep learning applications for predicting pharmacological properties of drugs and drug repurposing using transcriptomic data," *Molecular pharmaceutics,* vol. 13, no. 7, p. 2524–2530, 2016.

[84] H. Y. Zhang, Y. Lu and et al, "Development of a Survival Prognostic Model for Non-small Cell Lung Cancer," *Front Oncol,* vol. 10, p. 362, 2020.

[85] Z. Zhao, H. Peng and et al, "Identification of lung cancer gene markers through kernel maximum mean discrepancy and information entropy," *BMC Med Genomics,* vol. 12, no. Suppl 8, p. 183, 2019.

[86] J. A. Sim, Y. A. Kim and et al, "The major effects of health-related quality of life on 5-year survival prediction among lung cancer survivors: applications of machine learning," *Sci Rep,* vol. 10, no. 1, p. 10693, 2020.

[87] M. R. Salmanpour, M. Shamsaei and et al, "Optimized machine learning methods for prediction of cognitive outcome in Parkinson's disease," *Computers in Biology and Medicine,* vol. 111, p. 103347, 2019.

[88] M. R. Salmanpour, M. Shamsaei and et al, "Machine learning methods for optimal prediction of motor outcome in Parkinson's disease," *Physica Medica,* vol. 69, pp. 233-240, 2020.

[89] E. Nemlander, A. Rosenblad and et al, "Lung cancer prediction using machine learning on data from a symptom e-questionnaire for never smokers, formers smokers and current smokers," *PLoS ONE,* vol. 17, no. 10, p. e0276703, 2022.

[90] T. Y. Yang, P. Y. Kuo and et al, "Deep-Learning Approach to Predict Survival Outcomes Using Wearable Actigraphy Device Among End-Stage Cancer Patients," *Front Public Health,* vol. 9, p. 730150, 2021.

[91] S. Misra, R. Narayanan and et al, "Lung Cancer Survival Prediction using Ensemble Data Mining on Seer Data," *Scientific Programming,* vol. 20, p. 14, 2012.

[92] N. Khosravan and U. Bagci, "Semi-Supervised Multi-Task Learning for Lung Cancer Diagnosis," *Annu Int Conf IEEE Eng Med Biol Soc,* vol. 2018, pp. 710-713, 2018.

[93] P. Lambin, E. Rios-Velazquez and et al, "Radiomics: Extracting more information from medical images using advanced feature analysis," *European Journal of Cancer,* vol. 48, no. 4, pp. 441-446, 2012.

[94] J. Lao, Y. Chen and et al, "A Deep Learning-Based Radiomics Model for Prediction of Survival in Glioblastoma Multiforme," *Sci Rep,* vol. 7, no. 1, p. 10353, Sep 4 2017.